# Geometric and Topological Deep Learning for Predicting Thermo-mechanical Performance in Cold Spray Deposition Process Modeling

Akshansh Mishra[1]

[1]School of Industrial and Information Engineering, Politecnico di Milano, Milan, Italy

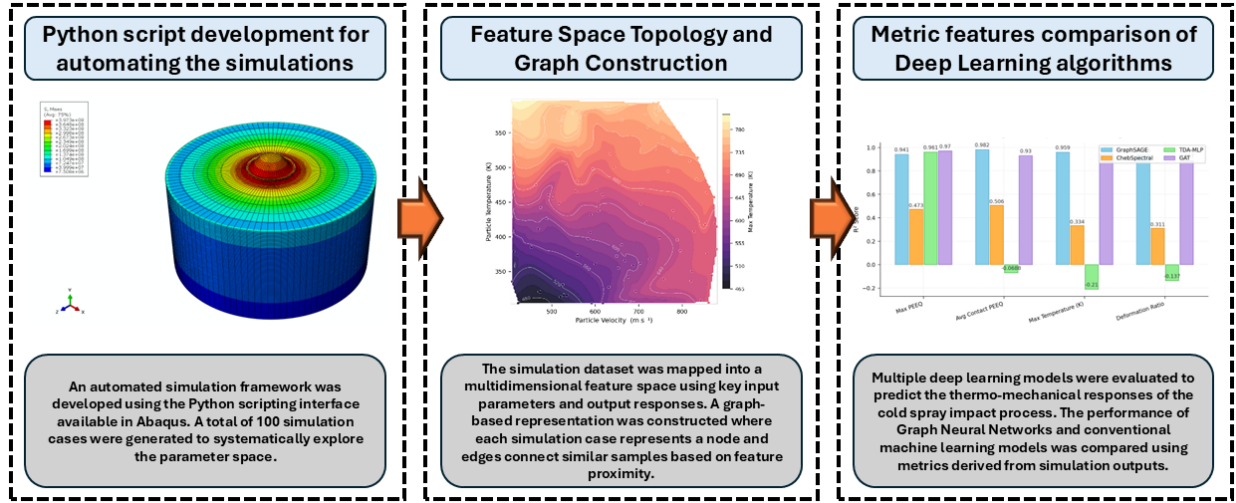

**Abstract:** This study presents a geometric deep learning framework for predicting cold spray particle impact responses using finite element simulation data. A parametric dataset was generated through automated Abaqus simulations spanning a systematic range of particle velocity, particle temperature, and friction coefficient, yielding five output targets including maximum equivalent plastic strain, average contact plastic strain, maximum temperature, maximum von Mises stress, and deformation ratio. Four novel algorithms i.e. a GraphSAGE-style inductive graph neural network, a Chebyshev spectral graph convolution network, a topological data analysis augmented multilayer perceptron, and a geometric attention network were implemented and evaluated. Each input sample was treated as a node in a k-nearest-neighbour feature-space graph, enabling the models to exploit spatial similarity between process conditions during training. Three-dimensional feature space visualisations and two-dimensional contour projections confirmed the highly non-linear and velocity-dominated nature of the input-output relationships. Quantitative evaluation demonstrated that GraphSAGE and GAT consistently achieved $R^2$ values exceeding 0.93 across most targets, with GAT attaining peak performance of $R^2 = 0.97$ for maximum plastic strain. ChebSpectral and TDA-MLP performed considerably worse, yielding negative $R^2$ values for several targets. These findings establish spatial graph-based neighbourhood aggregation as a robust and physically interpretable surrogate modelling strategy for cold spray process optimisation.

## 1. Introduction

Cold spray is a solid-state deposition process in which metallic particles are accelerated to supersonic velocities and directed toward a substrate, where bonding occurs through extreme plastic deformation at the particle-substrate interface rather than through melting or fusion [1-4]. Since its emergence as a practical coating and additive manufacturing technology, cold spray has attracted considerable research interest owing to its ability to deposit thermally sensitive materials without inducing phase transformations, oxidation, or residual porosity associated with thermal spray counterparts. The mechanical integrity of the deposited layer depends critically on the interfacial conditions established during the impact event, which are governed by a complex and highly coupled set of process parameters including particle impact velocity, particle temperature at the point of contact, and the tribological conditions prevailing at the interface. Understanding how these parameters interact to determine the resulting deformation state, stress field, and thermal response of the impacting particle remains a fundamental challenge in advancing cold spray towards precision-engineered applications in aerospace, biomedical, and energy sectors.

Finite element modelling has served as the principal investigative tool for resolving the transient mechanics of cold spray impact at the single-particle scale [5-7]. Early contributions established that bonding is facilitated by adiabatic shear instability at the particle periphery, where localised thermal softening momentarily overcomes strain hardening and promotes intimate metal-to-metal contact. Refined computational studies incorporated strain rate dependent constitutive models, cohesive zone formulations, and Arbitrary Lagrangian-Eulerian frameworks to better capture the extreme deformation gradients and jetting phenomena characteristic of successful particle adhesion. These advances notwithstanding, individual finite element simulations remain computationally expensive, and the parametric space defined by velocity, temperature, and friction is sufficiently large that exhaustive simulation-based exploration is practically infeasible. This has motivated growing interest in data-driven surrogate models capable of rapidly approximating the input-output relationships learned from a representative simulation dataset.

Machine learning approaches applied to cold spray and broader impact mechanics problems have progressed considerably in recent years. Regression-based methods, artificial neural networks, and ensemble learning algorithms have each demonstrated capacity to predict macroscopic deposition outcomes from process parameter inputs with reasonable accuracy [8-11]. However, these conventional architectures treat each simulation sample as an independent observation, discarding the geometric relationships that exist between samples occupying similar regions of the input parameter space. In physical systems where the response varies smoothly and continuously with the governing parameters, neighbouring conditions in feature space carry mutually informative context that a standard feedforward architecture cannot exploit. This representational limitation becomes particularly significant when training data is sparse and the underlying response surface is non-linear, as is characteristic of cold spray impact mechanics across the full operating envelope.

Geometric deep learning provides a principled framework for incorporating relational and structural information into the learning process by operating on graph-structured data representations rather than on independently treated feature vectors [12-15]. Graph neural networks propagate information across edges connecting similar nodes, enabling each sample to be informed by the responses of its nearest neighbours in the feature space and thereby capturing local geometric regularity that point-wise models cannot represent. Spectral approaches to graph convolution, including Chebyshev polynomial approximations of the graph Laplacian, extend this capability into the frequency domain and provide a theoretically grounded mechanism for multi-scale feature extraction on irregular domains. Topological data analysis offers a complementary perspective by encoding global shape properties of the data manifold through persistent homology, augmenting local neighbourhood information with descriptors sensitive to the connectivity and structure of the point cloud at multiple spatial scales.

The application of these methods to molecular property prediction, fluid dynamics surrogate modelling, and structural mechanics problems has demonstrated their broader utility in scientific computing contexts, yet their potential for cold spray process prediction has not been explored. The present work addresses this gap by constructing a parametric finite element simulation dataset spanning a systematic range of particle velocity, temperature, and friction coefficient, and by implementing and evaluating four geometric deep learning architectures on the resulting data. The study aims to establish which algorithmic strategy most effectively captures the non-linear, coupled input-output relationships characteristic of cold spray impact mechanics, and to demonstrate the value of graph-based representation as a physically motivated inductive bias for process-level surrogate modelling in solid-state deposition technologies.

## 2. Methodology

### 2.1. Geometry Definition used in the present work

Figure 1 shows the geometrical model used in the present work which shows the impact of a single metallic spherical particle onto a flat cylindrical surface. This adopted configuration is common for the investigation of cold spray deposition mechanisms. The proposed model consists of two primary components i.e. a spherical particle with a radius 40 μm and a cylindrical substrate of radius 250 μm and of depth 250 μm. The geometry of both these components is generated using the sketch and revolve approach within the Abaqus modeling environment.

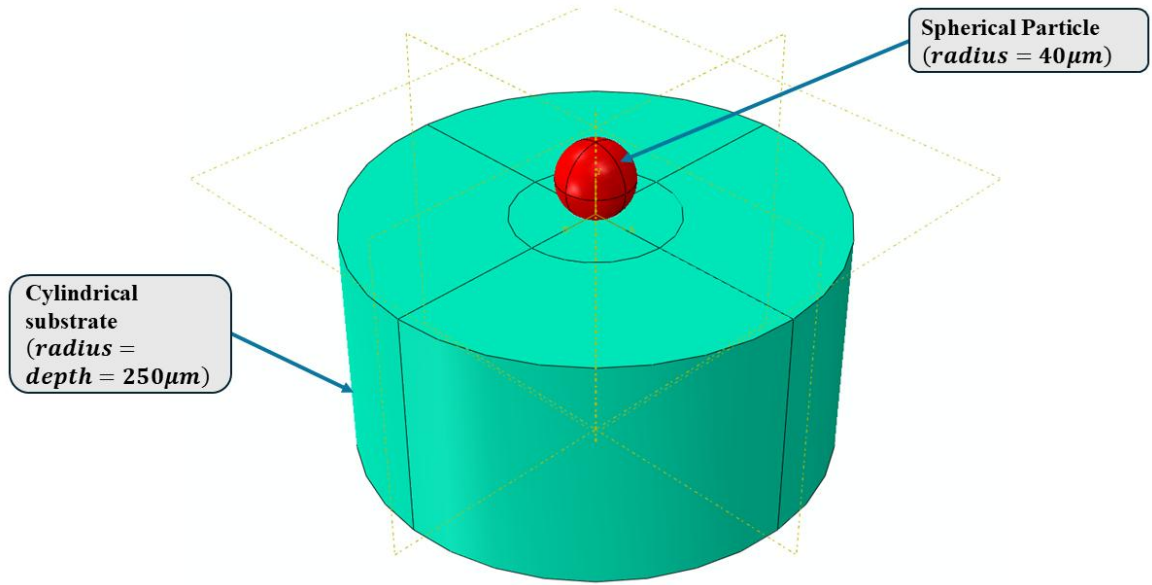

**Figure 1.** Geometrical configuration of the cold spray impact model used in the present work. The model consists of a spherical particle with a radius of 40 μm positioned above a cylindrical substrate with a radius and depth of 250 μm. The particle is aligned along the central axis and impacts the substrate in the vertical direction, representing a normal particle impact during the cold spray process.

The particle is positioned above the substrate surface in such a manner that its lower surface is aligned with the impact axis. The particle is translated vertically by a distance equal to its radius so that it is placed just above the substrate surface before the beginning of the impact phenomenon. The impact occurs along the vertical direction that represents the normal incidence of the spherical particle onto cylindrical the substrate surface, which is a typical assumption in single-particle cold spray impact studies.

### 2.2. Material Model used in the present work

In the present work both the substrate and the particle are assumed to be made of aluminum in order to represent a typical cold spray deposition process which involves aluminum particles impacting an aluminum substrate. The material behavior during the high-velocity impact process is modeled using a thermo-mechanically coupled constitutive framework that accounts for elastic deformation, plastic deformation, strain-rate effects, and temperature evolution. A rate-dependent plasticity model is necessary to accurately capture the material response due to the extremely high strain rates and severe plastic deformation experienced during particle impact.

The elastic behavior of the material is defined through its shear modulus and density, while thermal properties such as thermal conductivity and specific heat capacity are incorporated to represent heat generation and transfer during impact. Plastic deformation is modeled using the Johnson–Cook constitutive model, which is widely used for high strain-rate deformation processes including cold spray, ballistic impact, and explosive loading. This model accounts for strain hardening, strain-rate sensitivity, and thermal softening of the material. The strain-rate dependence of plastic deformation is incorporated through the Johnson–Cook rate-dependent formulation. The material model also includes an equation of state (EOS) based on the linear shock velocity–particle velocity (Us–Up) relationship to accurately capture pressure wave propagation and compressibility effects during high-velocity impact. This formulation allows the simulation to represent the pressure response of the material under extreme loading conditions. The conversion of plastic work into heat is also included through the inelastic heat fraction that enables thermo-mechanical coupling during deformation. Table 1 shows the material properties used in the present work.

Table 1. Material properties of aluminum used in the simulation

| Property | Symbol | Value | Unit |
| --- | --- | --- | --- |
| Density | $\rho$ | 2700 | kg m$^{-3}$ |
| Shear Modulus | $G$ | $27 \times 10^9$ | Pa |
| Thermal conductivity | $\kappa$ | 237.2 | W m$^{-1}$ K$^{-1}$ |
| Specific heat capacity | $C_p$ | 898.2 | J kg$^{-1}$ K$^{-1}$ |
| Johnson-Cook yield stress | $A$ | $148.4 \times 10^6$ | Pa |
| Johnson-Cook hardening constant | $B$ | $345.5 \times 10^6$ | Pa |
| Johnson-Cook hardening exponent | $n$ | 0.183 | - |
| Strain-rate coefficient | $C$ | 0.001 | - |
| Thermal softening exponent | $m$ | 0.895 | - |
| Melting temperature | $T_m$ | 916 | K |
| Reference temperature | $T_0$ | 298 | K |
| Bulk sound speed (EOS) | $C_0$ | 5386 | m s$^{-1}$ |
| Hugoniot slope coefficient | $S$ | 1.339 | - |
| Grüneisen parameter | $\Gamma_0$ | 1.97 | - |

**2.3. Meshing strategy used in the present work**

In the present work, both the spherical particle and the cylindrical substrate are discretized using three-dimensional thermo-mechanically coupled solid elements available in the Abaqus/Explicit solver. The element formulation includes reduced integration with temperature degrees of freedom to allow simultaneous solution of the mechanical and thermal fields during the impact event. The spherical particle is meshed using a relatively fine mesh in order to accurately capture the large plastic deformation and stress gradients that develop during impact. A global seed size of approximately 2.5 μm is applied to the particle domain. The substrate is meshed using a combination of coarse and refined mesh regions to balance computational efficiency and solution accuracy. A global mesh size of approximately 35 μm is applied to the bulk substrate region, while a significantly finer mesh of approximately 3.5 μm is used near the impact zone where high stress and strain gradients are expected.

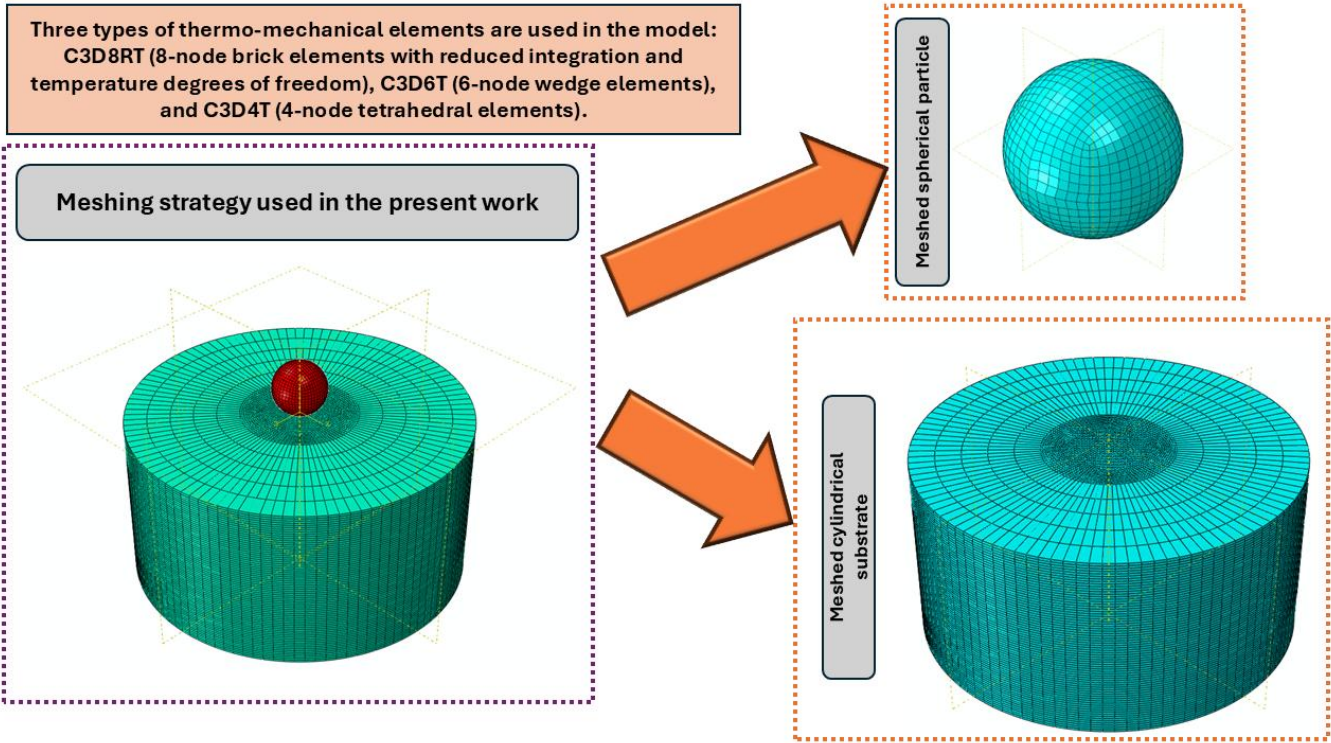

**Figure 2.** Meshing strategy adopted for the numerical model. The spherical particle and cylindrical substrate are discretized using thermo-mechanical elements, including C3D8RT (8-node brick elements with reduced integration and temperature degrees of freedom), C3D6T (6-node wedge elements), and C3D4T (4-node tetrahedral elements). A finer mesh is applied near the particle–substrate contact region to accurately capture high stress, strain, and temperature gradients during impact, while a relatively coarser mesh is used away from the impact zone to improve computational efficiency.

Three types of thermo-mechanical elements are used in the model: C3D8RT (8-node brick elements with reduced integration and temperature degrees of freedom), C3D6T (6-node wedge elements), and C3D4T (4-node tetrahedral elements). The C3D8RT elements are primarily used in regions where structured meshing is possible, while wedge and tetrahedral elements are employed to accommodate geometric transitions and maintain mesh compatibility.

An adaptive mesh refinement technique based on the Arbitrary Lagrangian–Eulerian (ALE) formulation is applied to both the particle and the region of the substrate near the impact zone is used to further improve numerical stability and handle the extreme deformation occurring during particle impact. The adaptive meshing procedure allows the mesh to continuously adjust during the simulation, reducing element distortion and maintaining element quality throughout the high-deformation process. This approach is particularly important in cold spray simulations where large plastic deformation and severe contact interactions can otherwise lead to mesh distortion and premature termination of the simulation.

### 2.4. Automated Simulation Framework, Parametric Study and Data Extraction

An automated simulation framework shown in Figure 3 was developed using the Python scripting interface available in Abaqus to efficiently perform a large number of cold spray impact simulations. The framework allows the entire simulation workflow including model generation, parameter assignment, job execution, and post-processing to be performed automatically without manual intervention.

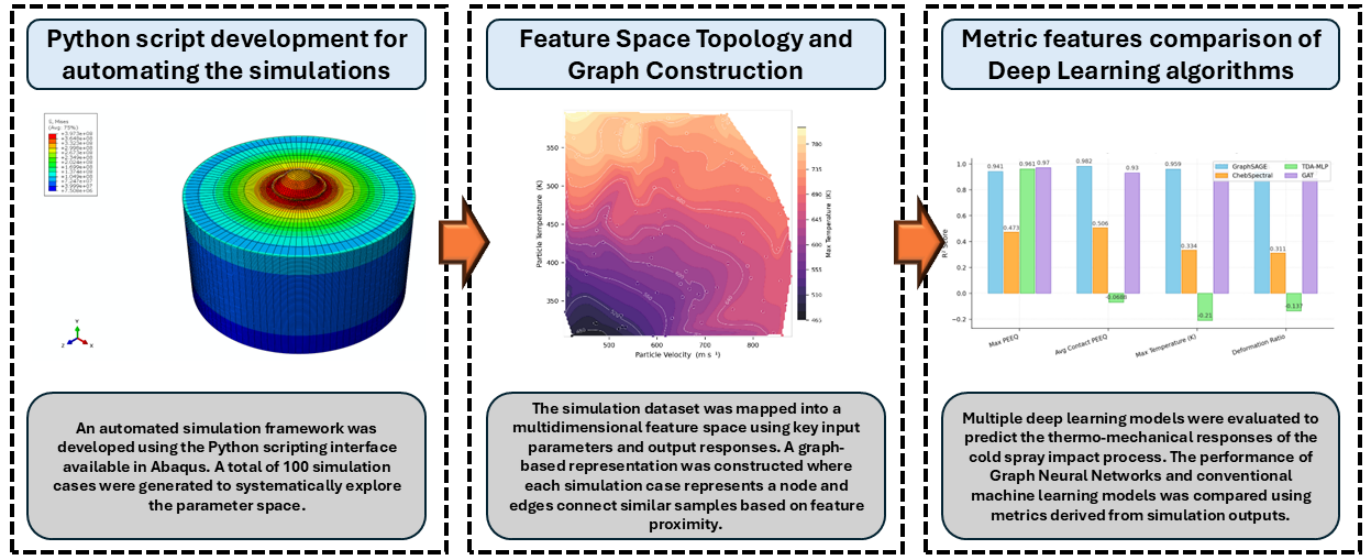

**Figure 3.** The simulation–deep learning framework used in this study. An automated Python scripting framework in Abaqus was developed to generate a dataset of cold spray particle impact simulations across the selected parameter space. The resulting simulation data were mapped into a multidimensional feature space using key input parameters and output responses, and a graph-based representation was constructed where nodes represent simulation cases and edges capture similarity relationships between samples. Multiple deep learning models were then evaluated using metrics derived from the simulation outputs to assess their capability in predicting the thermo-mechanical responses of the cold spray impact process.

The simulation process begins with the definition of a set of input parameters representing different cold spray conditions. In the present work, three key parameters i.e. particle impact velocity, particle temperature, and the coefficient of friction at the particle–substrate interface are varied as shown in Table 2. These parameters play an important role in determining the deformation behavior, bonding mechanism, and thermal response during the cold spray process. A total of 100 simulation cases were generated to systematically explore the parameter space.

**Table 2.** Input parameters used in the parametric simulation study

| Parameter | Symbol | Range | Unit | Description |
|---|---|---|---|---|
| Particle velocity | $V$ | 400-900 | m s$^{-1}$ | Initial velocity applied to the particle |
| Particle temperature | $T_p$ | 300-600 | K | Initial temperature of the particle before impact |
| Friction coefficient | $\mu$ | 0.10-0.50 | - | Coulomb friction coefficient at the particle–substrate interface |

Several performance indicators are calculated to characterize the particle impact behavior. These include the maximum equivalent plastic strain (PEEQ), average plastic strain in the contact region, maximum temperature reached during impact, maximum von Mises stress, and the deformation ratio of the particle. The deformation ratio is used as a measure of particle flattening and is calculated based on the maximum downward displacement of the particle relative to its original diameter. All extracted results, along with the corresponding input parameters, are stored in a structured dataset for further analysis.

## 2.5. Geometric Deep Learning algorithms used in the present work

Four data-driven models i.e. GraphSAGE, Chebyshev Spectral GNN, TDA-augmented MLP, and Geometric Attention Network (GAT) were implemented. The input feature vector for each sample is defined as $x_i \epsilon \mathbb{R}^d$ where the features correspond to the particle velocity, particle temperature, and friction coefficient. A k-nearest neighbor (k-NN) graph was constructed in the normalized feature space where each simulation case is treated as a node and edges connect neighbouring samples. The edge weights were computed using a Gaussian kernel using Equation 1.

$$w_{ij} = exp\left(-\frac{d^2_{ij}}{2\sigma^2}\right) \tag{1}$$

Where $d_{ij}$ is the Euclidean distance between nodes $i$ and $j$, and $\sigma$ is a scale parameter derived from the median edge distance.

The node representations are updated by combining information from neighbouring nodes through weighted mean message passing in the GraphSAGE model. The hidden representation of node $i$ at layer $l + 1$ is computed using Equation 2.

$$h_i^{(l+1)} = \sigma\left(W_s^{(l)} h_i^{(l)} + W_n^{(l)} AGG(\{h_j^{(l)}: j \epsilon \aleph(i)\})\right) \tag{2}$$

Where $h_i^{(l)}$ is the node embedding at layer $l$, $\aleph(i)$ denotes the neighbourhood of node $i$, $W_s^{(l)}$ and $W_n^{(l)}$ are trainable weight metrics, and $\sigma(.)$ is a nonlinear activation function. This model captures the local geometric relationships in the feature-space graph.

The Chebyshev Spectral GNN performs graph convolution in the spectral domain using Chebyshev polynomial approximation of the graph Laplacian. The convolution operation is computed using Equation 3.

$$g_\theta(L)x = \sum_{k=0}^{K-1} \theta_k T_k(L) x \tag{3}$$

Where $L$ is the normalized graph Laplacian, $T_k(L)$ is the Chebyshev polynomial of order $k$, and $\theta_k$ are learnable coefficients. This formulation allows efficient multi-scale feature propagation over the graph without requiring eigen decomposition of the Laplacian.

The original input features are enriched by topological descriptors derived from pairwise-distance structure in the dataset in the TDA-augmented MLP. These descriptors include connectivity and density-related proxies inspired by persistent homology, and the augmented feature vector is then passed through a multilayer perceptron computed using Equation 4.

$$h^{(l+1)} = \emptyset\left(W^{(l)} h^{(l)} + b^{(l)}\right) \tag{4}$$

Where $W^{(l)}$ and $b^{(l)}$ are the learnable weights and biases, and $\emptyset(.)$ denotes the nonlinear activation. This approach combines raw process parameters with topology-aware descriptors to improve predictive capability.

The Geometric Attention Network (GAT) uses an attention mechanism to assign different importance to neighbouring nodes during message passing. The attention coefficient between connected nodes $i$ and $j$ is computed using Equation 5.

$$\alpha_{ij} = \frac{exp\left(LeakyReLU(a^T[Wh_i || Wh_j])\right)}{\sum_{k \in \aleph(i)} exp\left(LeakyReLU(a^T[Wh_i || Wh_k])\right)} \tag{5}$$

Where $W$ is a learnable transformation matrix, $a$ is the attention vector, and || denotes concatenation. In the present implementation, these attention scores are further modulated by edge weights derived from geometric distances, allowing the network to emphasize more relevant neighbours in the feature-space graph. The models were trained to minimize the mean squared error (MSE) between predicted and actual target values using Equation 6.

$$MSE = \frac{1}{N}\sum_{i=1}^{N}(y_i - \hat{y}_i)^2 \qquad (6)$$

and their predictive performance was assessed using the coefficient of determination $R^2$, mean absolute error (MAE), and MSE. This framework enables comparison between graph-based and topology-enhanced learning methods for predicting thermo-mechanical responses in the cold spray process.

**3. Results and Discussion**

**3.1. Finite Element Analysis results**

Figure 4a presents the von Mises stress distribution across the particle and substrate, revealing a highly concentrated stress field at the impact interface that radiates outward in a characteristic butterfly pattern. The cross-sectional views confirm that peak stress is localised within the deformed particle and the immediate subsurface region, with the surrounding substrate retaining comparatively low stress values, consistent with the highly localised nature of cold spray impact mechanics.

Figure 4b illustrates the equivalent plastic strain (PEEQ) distribution, where severe plastic deformation is confined almost entirely to the particle and a narrow contact zone at the substrate surface. The majority of the substrate volume remains at near-zero PEEQ, demonstrating that plastic deformation does not propagate deeply into the substrate and that bonding is governed by the extreme localised straining at the particle periphery rather than bulk material yielding.

Figure 4c shows the nodal temperature field at peak impact, with elevated temperatures restricted to a thin interfacial layer between the particle and substrate. The thermal gradient decays sharply away from the contact zone in all directions, indicating that adiabatic shear heating is highly localised and transient, supporting the established understanding that thermal softening at the interface, rather than bulk melting, is the primary thermally driven mechanism facilitating particle adhesion in cold spray deposition.

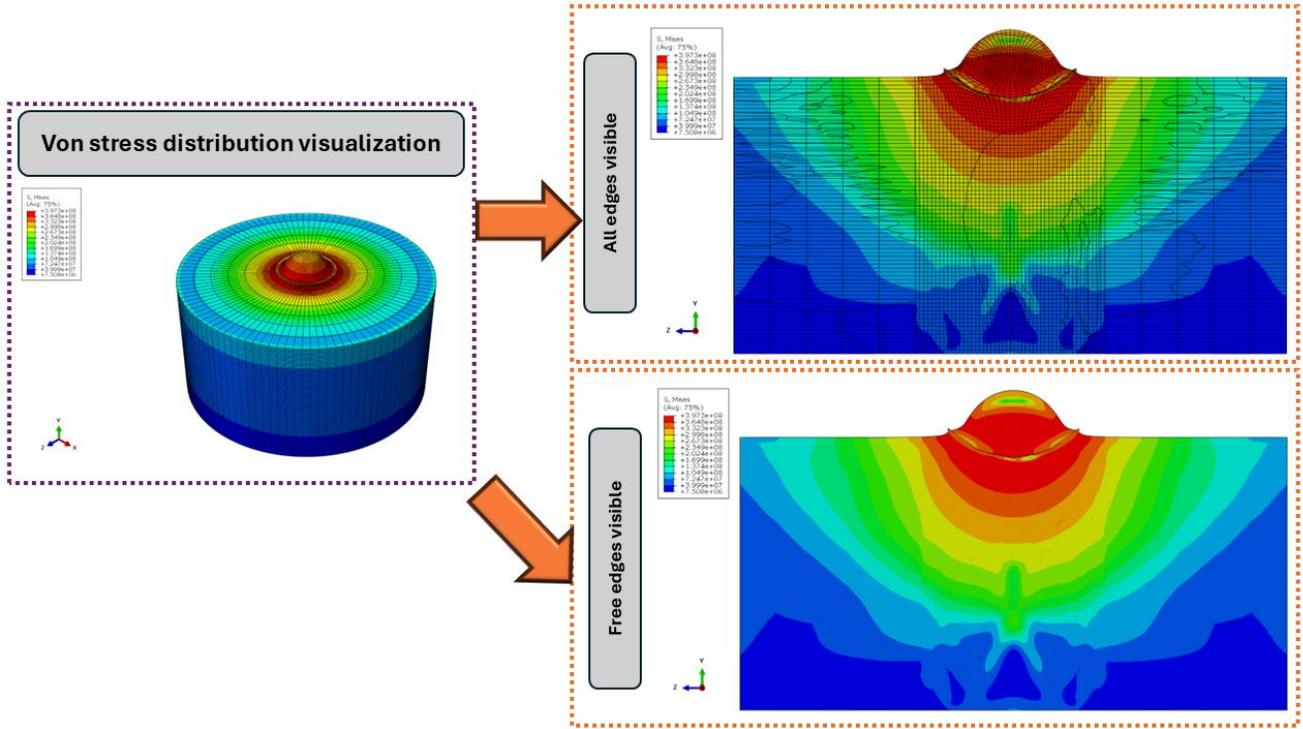

a)

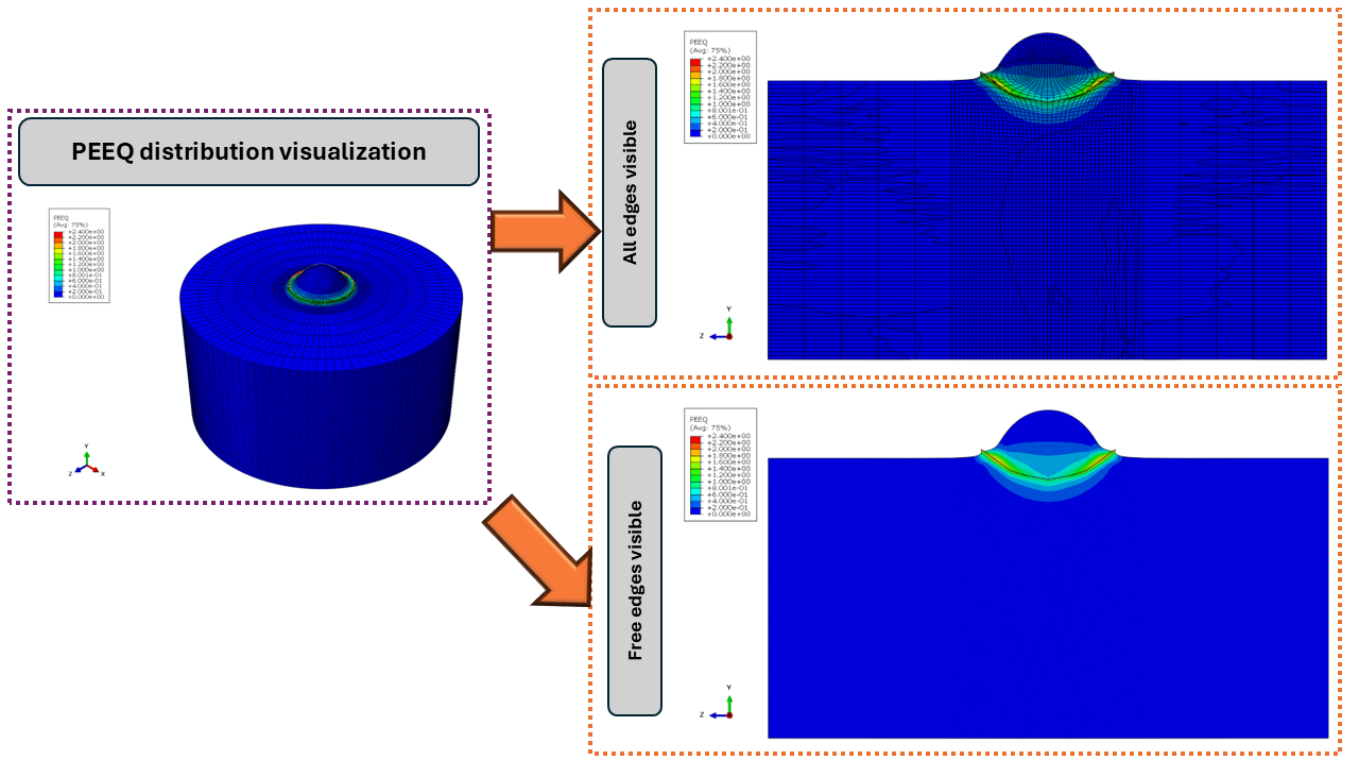

b)

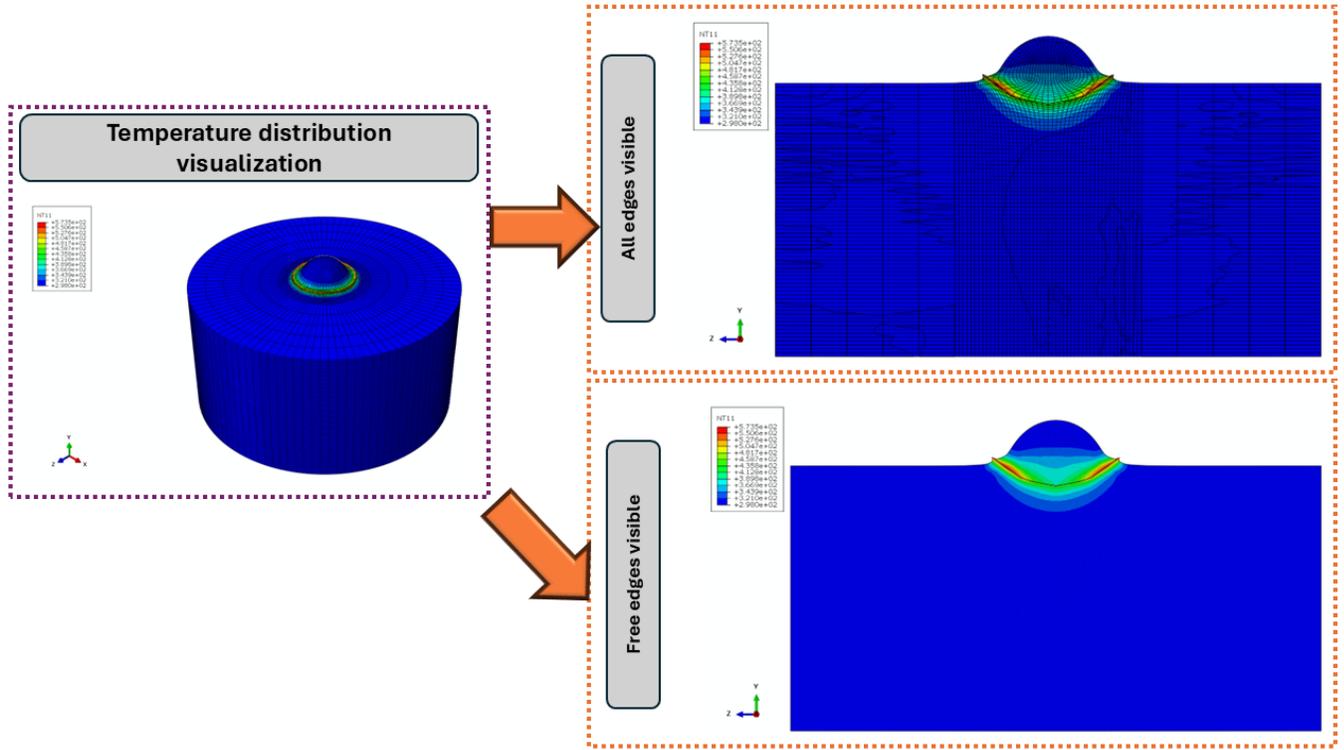

c)

**Figure 4.** Finite element simulation contour plots of the cold spray impact event showing cross-sectional distributions under all-edges-visible and free-edges-visible display conditions: (a) von Mises stress distribution revealing a concentrated stress field at the particle-substrate interface that propagates outward in a butterfly-shaped pattern into the substrate; (b) equivalent plastic strain (PEEQ) distribution confirming that severe plastic deformation is localised within the impacting particle and the immediate contact zone, with the surrounding substrate remaining largely undeformed; and (c) nodal temperature distribution demonstrating that adiabatic shear-induced heating is confined to a thin interfacial layer at the point of contact, decaying sharply in all directions.

## 3.2. Feature Space Topology and Graph Construction

Figure 5a (Velocity × Temperature) presents the clearest response, with PEEQ increasing sharply and monotonically beyond ~650 m s$^{-1}$, while temperature acts as a secondary amplifier at higher velocities. The smooth, well-resolved gradient confirms that these two parameters jointly constitute the primary drivers of plastic deformation in cold spray. Figure 5b (Velocity × Friction Coefficient) reveals strong vertical banding, indicating that PEEQ is governed almost exclusively by particle velocity irrespective of the friction coefficient, with contour lines running nearly parallel to the friction axis. Isolated high-value points at elevated velocities suggest that friction can transiently amplify deformation under specific conditions, though this effect lacks consistency across the dataset. Figure 5c (Temperature × Friction Coefficient) exhibits the most irregular contour morphology, with a notably wide colorbar range that includes interpolation artefacts below zero (−4.5), reflecting sparse data coverage once velocity is projected out. This panel demonstrates that neither temperature nor friction coefficient, in isolation, carries sufficient explanatory power for PEEQ, reinforcing the necessity of capturing all three input dimensions simultaneously.

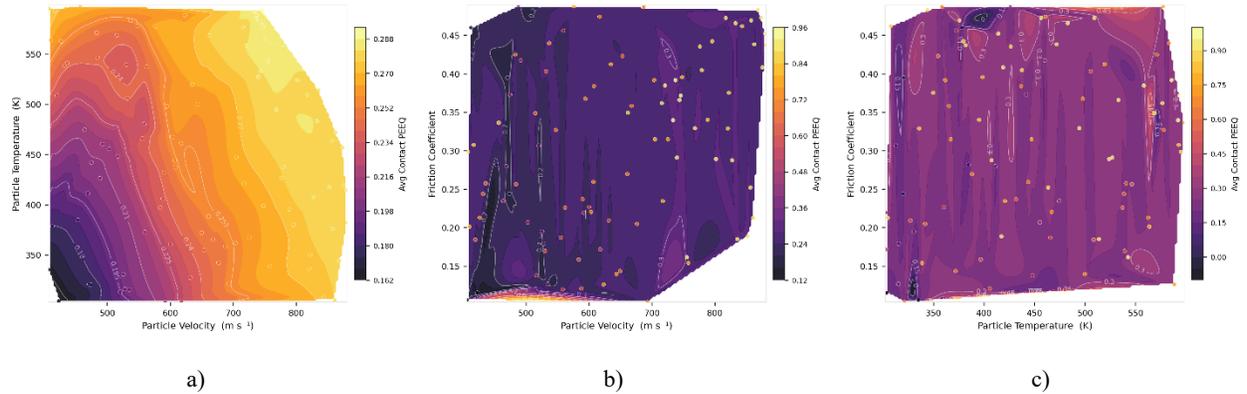

**Figure 5.** Two-dimensional contour projections of maximum plastic equivalent strain (PEEQ) across the three pairwise input combinations: (a) particle velocity versus particle temperature, (b) particle velocity versus friction coefficient, and (c) particle temperature versus friction coefficient. Filled contours represent the interpolated PEEQ response surface (cubic griddata), white isolines denote constant-strain boundaries, and overlaid scatter points correspond to actual simulation samples coloured by their PEEQ value. The progressive loss of contour regularity from (a) to (c) reflects the diminishing explanatory power of each input pair when particle velocity is excluded, highlighting the inherently three-dimensional, non-linear nature of the deformation response.

Figure 6a (Friction ≈ 0.31, Velocity × Temperature slice) displays the smoothest and most physically interpretable surface, forming a clear saddle-like geometry where average contact PEEQ rises with increasing velocity and falls toward intermediate temperatures (~400–450 K), before recovering at higher thermal values. The regularity of this surface, with minimal scatter deviation from the interpolated mesh, confirms that at a fixed friction coefficient the velocity–temperature interaction governs contact deformation in a well-structured, predictable manner. Figure 6b (Temperature ≈ 442 K, Velocity × Friction slice) reveals a generally ascending surface with velocity as the dominant axis, but introduces modest undulation along the friction direction at high velocities (>750 m s$^{-1}$). The surface remains largely coherent, though the trailing edge near low friction and high velocity begins to show slight irregularity, suggesting that at elevated kinetic energy, friction coefficient starts to modulate contact deformation non-trivially. Figure 6c (Velocity ≈ 623 m s$^{-1}$, Temperature × Friction slice) is the most topographically complex panel, exhibiting pronounced ridges, local minima, and surface corrugation across the temperature–friction plane. This roughness is consistent with sparse sampling in this projection once velocity is held constant, but also physically indicates that the combined thermal and tribological conditions produce competing deformation mechanisms at intermediate velocities, neither fully kinetic nor fully thermally driven.

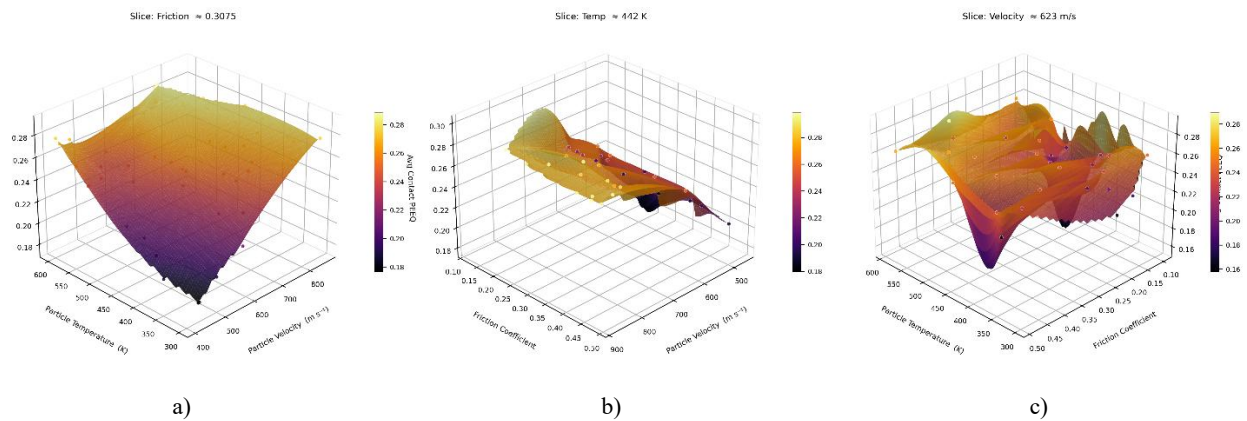

**Figure 5.** Three-dimensional response surfaces of average contact plastic equivalent strain (Avg Contact PEEQ) obtained by fixing one input parameter near its median value and sweeping the remaining two: (a) friction coefficient fixed at approximately 0.31, sweeping particle velocity and temperature; (b) particle temperature fixed at approximately 442 K, sweeping particle

velocity and friction coefficient; and (c) particle velocity fixed at approximately 623 m s$^{-1}$, sweeping particle temperature and friction coefficient. Surfaces are generated via cubic interpolation on a structured meshgrid, with overlaid scatter points representing actual simulation samples. The progressive increase in surface corrugation from (a) to (c) reflects the diminishing smoothness of the response once particle velocity, the dominant input parameter, is progressively constrained, revealing the coupled, non-linear dependence of contact deformation on the full three-dimensional input space.

Figure 7a (Velocity × Temperature) presents the clearest and most physically interpretable response, with PEEQ increasing sharply and monotonically beyond approximately 650 m s$^{-1}$ while particle temperature acts as a secondary amplifier at higher velocities. The smooth, well-resolved gradient and tightly spaced isolines in the high-velocity region confirm that these two parameters jointly constitute the primary drivers of maximum plastic deformation in cold spray, with the kinetic contribution of velocity dominating over the thermal contribution across the entire parameter range. Figure 7b (Velocity × Friction Coefficient) reveals pronounced vertical banding throughout the contour field, indicating that maximum PEEQ is governed almost exclusively by particle velocity irrespective of the friction coefficient, with isolines running nearly parallel to the friction axis. Isolated high-value scatter points concentrated at elevated velocities suggest that friction can transiently amplify deformation under specific localised conditions, though this effect lacks systematic consistency across the dataset and does not manifest as a coherent gradient in the interpolated surface. Figure 7c (Temperature × Friction Coefficient) exhibits the most irregular contour morphology of the three projections, with a wide colorbar range that includes physically implausible interpolation artefacts below zero, reflecting sparse and unevenly distributed data coverage once velocity is projected out. This panel demonstrates that neither particle temperature nor friction coefficient, considered in isolation from velocity, carries sufficient explanatory power for maximum PEEQ, reinforcing the necessity of capturing all three input dimensions simultaneously and underscoring the motivation for a non-linear, graph-based modelling approach that operates across the full feature space.

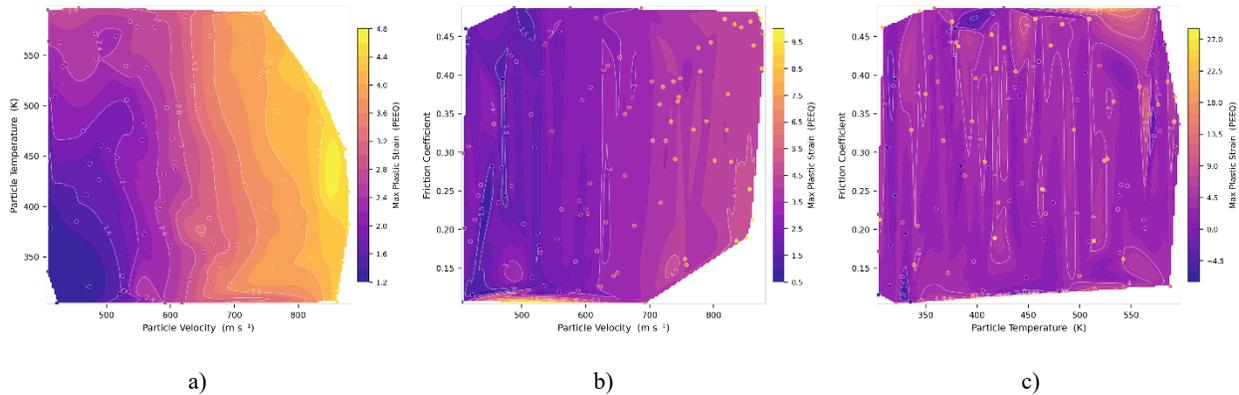

a)            b)            c)

**Figure 7.** Two-dimensional contour projections of maximum plastic equivalent strain (PEEQ) across the three pairwise input combinations: (a) particle velocity versus particle temperature, (b) particle velocity versus friction coefficient, and (c) particle temperature versus friction coefficient. Filled contours represent the interpolated PEEQ response surface obtained via cubic griddata interpolation, white isolines denote constant strain boundaries, and overlaid scatter points correspond to actual simulation samples coloured by their respective PEEQ values. The progressive loss of contour regularity from (a) to (c) reflects the diminishing explanatory power of each input pair when particle velocity is excluded, highlighting the inherently three-dimensional and non-linear nature of the plastic deformation response and motivating the use of a graph-based learning framework capable of resolving the full input space simultaneously.

Figure 8a (Friction ≈ 0.31, Velocity × Temperature slice) produces the smoothest and most well-behaved surface, displaying a near-planar monotonic rise in maximum PEEQ with increasing particle velocity, while temperature contributes a gentle modulating gradient along the orthogonal axis. The close agreement between the interpolated

mesh and the overlaid scatter points in this panel confirms that at a fixed friction coefficient, the velocity and temperature inputs jointly define a coherent, predictable deformation landscape with minimal unexplained variance. Figure 8b (Temperature ≈ 442 K, Velocity × Friction slice) shows a surface that ascends steeply along the velocity axis while remaining comparatively flat across the friction dimension at low to moderate velocities. Beyond approximately 750 m s$^{-1}$, the trailing edge of the surface begins to curve and narrow, suggesting that the available data becomes sparse at high kinetic energy combined with low friction values, and that the friction coefficient begins to exert a non-trivial secondary influence only under these extreme conditions. Figure 8c (Velocity ≈ 623 m s$^{-1}$, Temperature × Friction slice) is the most topographically complex panel, characterised by sharp ridges, deep local minima, and significant surface corrugation distributed across the temperature and friction plane. This irregular geometry reflects both the sparse data density in this projection and the physically competing deformation mechanisms that emerge at intermediate velocities, where neither kinetic nor thermal driving forces fully dominate, resulting in a sensitive and highly localised response to changes in temperature and friction coefficient.

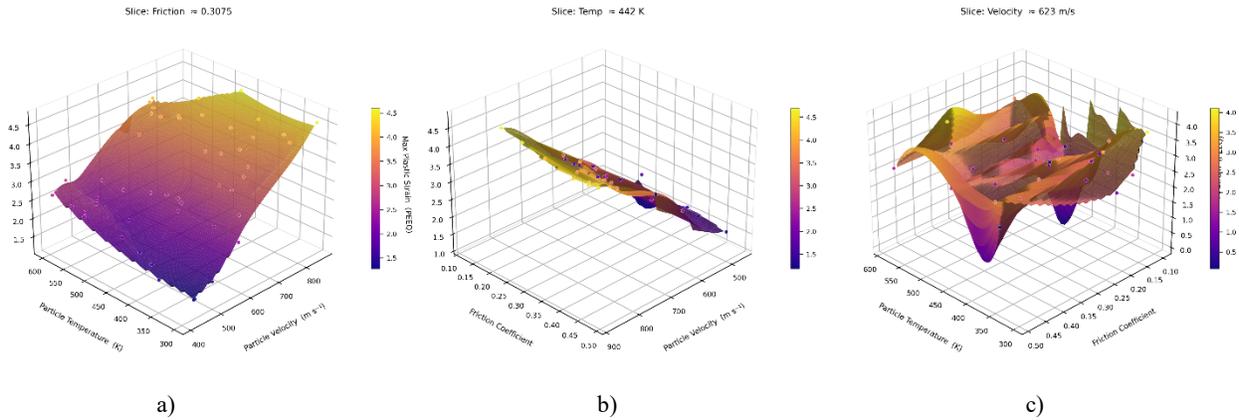

a)　　　　　　　　　　　　　　　　b)　　　　　　　　　　　　　　　　c)

**Figure 8.** Three-dimensional response surfaces of maximum plastic equivalent strain (PEEQ) obtained by fixing one input parameter near its median value and sweeping the remaining two: (a) friction coefficient fixed at approximately 0.31, sweeping particle velocity and temperature; (b) particle temperature fixed at approximately 442 K, sweeping particle velocity and friction coefficient; and (c) particle velocity fixed at approximately 623 m s$^{-1}$, sweeping particle temperature and friction coefficient. Surfaces are generated via cubic interpolation on a structured meshgrid, with overlaid scatter points representing actual simulation samples coloured by their corresponding PEEQ values. The progressive transition from a smooth, monotonically ascending surface in (a) toward an increasingly corrugated and locally irregular geometry in (c) reflects the diminishing regularity of the maximum plastic strain response once particle velocity, the dominant process parameter, is held constant, exposing the complex coupled interactions between particle temperature and friction coefficient at intermediate kinetic energy conditions.

Figure 9a (Velocity × Temperature) presents a broadly smooth and gradually ascending response, with maximum temperature rising consistently as both particle velocity and particle temperature increase. The contour gradient is relatively uniform across the parameter space, though a subtle undulation near the mid-velocity range (approximately 550 to 650 m s$^{-1}$) suggests a transitional regime where the thermal response becomes less linearly coupled to the inputs. The overall regularity of this panel confirms that the velocity and temperature inputs together provide a physically coherent and well-conditioned description of the maximum thermal field generated during impact. Figure 9b (Velocity × Friction Coefficient) displays the most severe irregularity of the three projections, with a colorbar range extending to approximately 2400 K and deep vertical channels of anomalously low interpolated values interspersed among high-temperature scatter points. This extreme spread and the erratic contour topology are indicative of strong interpolation instability driven by sparse and unevenly distributed data in this projection plane, rather than a genuine physical response. The scatter points themselves are broadly consistent in colour, suggesting the underlying data is physically reasonable but insufficiently dense to support stable surface reconstruction when velocity and friction are considered in isolation. Figure 9c (Temperature × Friction Coefficient) similarly exhibits vertical banding and a wide colorbar range, though with somewhat less extreme artefacting than Figure 9b. The dominant colour field remains uniformly dark across most of the projection, with elevated temperature values concentrated at the boundaries, reflecting that maximum temperature is weakly resolved by the temperature and friction inputs alone once the kinetic contribution of velocity is removed.

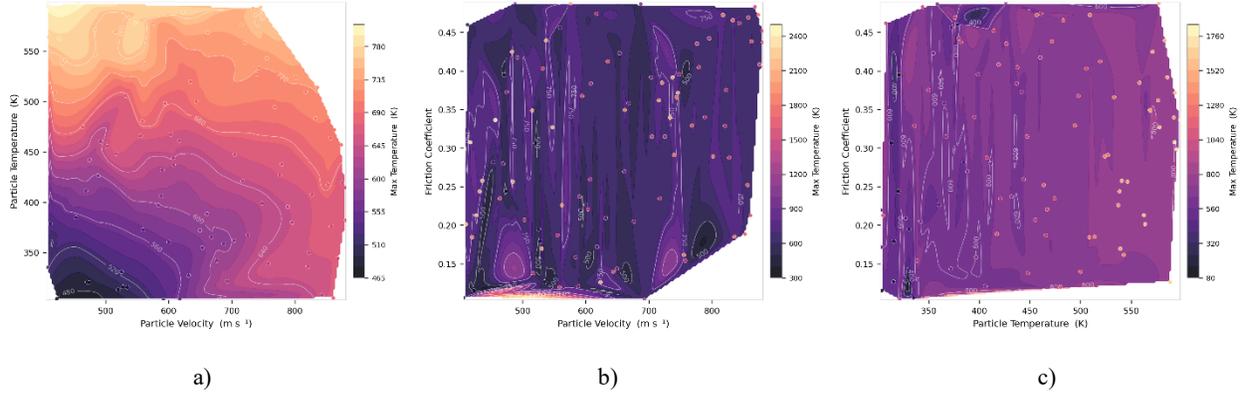

a)  b)  c)

**Figure 9.** Two-dimensional contour projections of maximum temperature (K) across the three pairwise input combinations: (a) particle velocity versus particle temperature, (b) particle velocity versus friction coefficient, and (c) particle temperature versus friction coefficient. Filled contours represent the interpolated maximum temperature response surface obtained via cubic griddata interpolation, white isolines denote constant temperature boundaries, and overlaid scatter points correspond to actual simulation samples coloured by their respective maximum temperature values. The well-structured gradient observed in (a) progressively deteriorates in (b) and (c) as particle velocity is excluded from the projection plane, with the colorbar range expanding substantially due to interpolation instability in data-sparse regions, confirming that maximum impact temperature is principally a velocity-governed output whose response topology is only faithfully captured when all three input dimensions are considered simultaneously.

Figure 10a (Friction ≈ 0.31, Velocity × Temperature slice) yields the most regular and physically interpretable surface of the three panels, displaying a smooth monotonic ascent in maximum temperature with increasing particle velocity while particle temperature contributes a secondary additive gradient along the orthogonal axis. The close correspondence between the interpolated mesh and the overlaid scatter points across the entire sweep range confirms that at a fixed friction coefficient, the thermal response field is well-conditioned and governed in a straightforward, predictable manner by the combined kinetic and thermal inputs. Figure 10b (Temperature ≈ 442 K, Velocity × Friction slice) introduces considerable surface complexity, with multiple undulations, localised ridges, and depressed valleys distributed across the velocity and friction plane. While the general trend of increasing maximum temperature with velocity is preserved, the surface morphology is disrupted by pronounced oscillations particularly at intermediate velocities between approximately 550 and 700 m s$^{-1}$, suggesting that at a fixed particle temperature, the friction coefficient interacts with velocity in a spatially inconsistent manner that the cubic interpolation struggles to resolve with the available data density. Figure 10c (Velocity ≈ 623 m s$^{-1}$, Temperature × Friction slice) exhibits the most severe surface degradation of the three panels, characterised by sharp discontinuous ridges, abrupt step-like transitions, and regions of near-vertical surface folding across the temperature and friction plane. This behaviour reflects the near-complete absence of a smooth underlying trend when velocity is held constant at an intermediate value, with the thermal response becoming highly sensitive to small variations in both remaining inputs and effectively unresolvable through interpolation alone at this level of data sparsity.

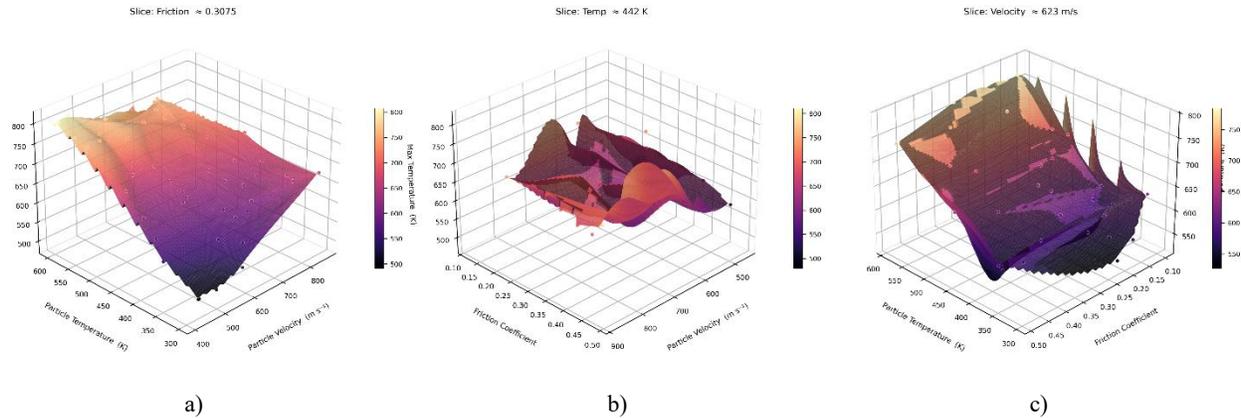

| a) | b) | c) |

**Figure 10.** Three-dimensional response surfaces of maximum temperature (K) obtained by fixing one input parameter near its median value and sweeping the remaining two: (a) friction coefficient fixed at approximately 0.31, sweeping particle velocity and particle temperature; (b) particle temperature fixed at approximately 442 K, sweeping particle velocity and friction coefficient; and (c) particle velocity fixed at approximately 623 m s$^{-1}$, sweeping particle temperature and friction coefficient. Surfaces are generated via cubic interpolation on a structured meshgrid, with overlaid scatter points representing actual simulation samples coloured by their corresponding maximum temperature values. The smooth monotonic surface observed in (a) progressively gives way to increasingly undulated and topographically irregular geometries in (b) and (c), reflecting the growing instability of the interpolated thermal response as particle velocity, the dominant governing parameter, is removed from the sweep space and the secondary coupled interactions between particle temperature and friction coefficient are foregrounded.

Figure 11a (Velocity × Temperature) reveals a spatially complex and non-monotonic response, with maximum von Mises stress exhibiting localised high-value concentrations near the mid-velocity range (approximately 580 to 650 m s$^{-1}$) at elevated particle temperatures around 550 K, rather than increasing uniformly with either input. The presence of closed contour loops and alternating high and low stress pockets across the parameter space indicates that the von Mises stress response is governed by competing mechanisms, where increasing velocity simultaneously drives deformation and softening, producing a non-trivial landscape that resists simple parametric description. Figure 11b (Velocity × Friction Coefficient) displays moderate vertical banding consistent with velocity dominance, but with considerably more lateral variation than observed in the equivalent PEEQ projections, suggesting that friction coefficient exerts a more meaningful secondary influence on stress than on plastic strain. A narrow band of elevated stress values is visible near 580 to 620 m s$^{-1}$ across multiple friction levels, reinforcing the existence of a transitional velocity regime where the stress state is particularly sensitive to the prevailing tribological conditions at the contact interface. Figure 11c (Temperature × Friction Coefficient) presents the broadest and most uniformly distributed colorbar range of the three panels, spanning approximately 195 to 435 MPa, with large contiguous regions of intermediate stress values and scattered localised extremes. The relatively diffuse and structurally weak contour pattern in this panel, compared to the corresponding PEEQ and temperature projections, suggests that von Mises stress has a more distributed sensitivity across all three inputs and is less dominated by a single parameter, making it the most challenging output to resolve through any two-variable projection alone.

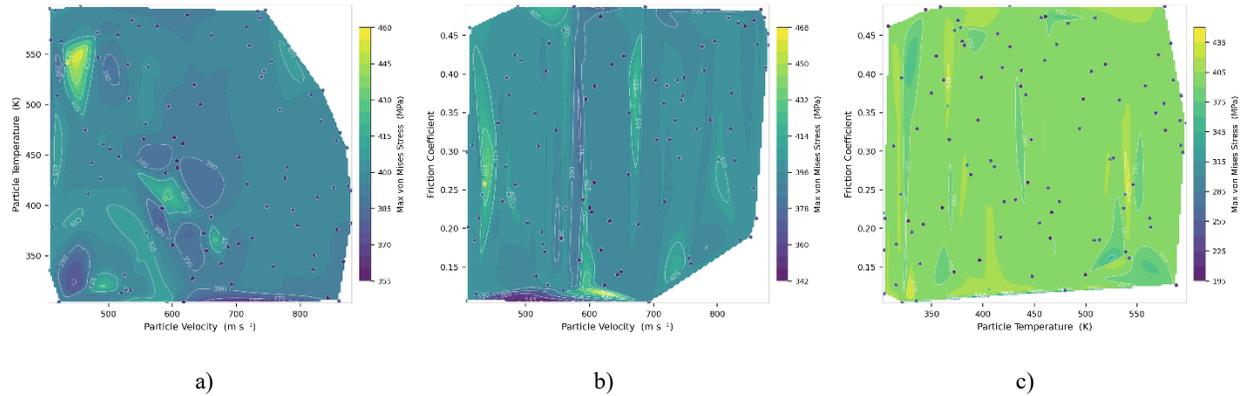

a)                                           b)                                           c)

**Figure 11.** Two-dimensional contour projections of maximum von Mises stress (MPa) across the three pairwise input combinations: (a) particle velocity versus particle temperature, (b) particle velocity versus friction coefficient, and (c) particle temperature versus friction coefficient. Filled contours represent the interpolated stress response surface obtained via cubic griddata interpolation, white isolines denote constant stress boundaries, and overlaid scatter points correspond to actual simulation samples coloured by their respective von Mises stress values.

Figure 12a (Friction ≈ 0.31, Velocity × Temperature slice) immediately distinguishes itself from the equivalent surfaces of other output targets by the presence of a sharp, localised stress peak rising prominently above an otherwise comparatively flat plateau at approximately 420 MPa. This isolated spike, concentrated near intermediate particle temperatures around 450 to 500 K at lower velocities, is indicative of a highly localised stress concentration that emerges under specific thermal and kinetic conditions rather than as part of a smooth monotonic trend, and its isolation from the surrounding surface suggests it corresponds to a narrow parametric regime where deformation hardening momentarily outpaces thermal softening. Figure 12b (Temperature ≈ 442 K, Velocity × Friction slice) presents the most dramatic surface geometry of the three panels, dominated by a pronounced central peak reaching approximately 540 MPa at intermediate velocities near 600 m s$^{-1}$ and low friction coefficients around 0.15 to 0.20, which then descends steeply in all directions. This sharp apex and the surrounding valley structure confirm that at a fixed particle temperature, the combined velocity and friction space contains a well-defined stress maximum that is highly sensitive to small perturbations in either input, a behaviour consistent with a transitional impact regime where the material briefly sustains maximum elastic stress before yielding. Figure 12c (Velocity ≈ 623 m s$^{-1}$, Temperature × Friction slice) displays a richly corrugated surface with multiple competing peaks, ridges, and depressions distributed across the temperature and friction plane, with stress values spanning the full colorbar range from approximately 370 to 430 MPa within relatively small regions of the input space. The dense and irregular topography of this panel, more complex than any of the equivalent slices for other targets, confirms that at intermediate velocities the von Mises stress response is exquisitely sensitive to the precise combination of particle temperature and friction coefficient, with no dominant gradient direction and no smooth transition between high and low stress regions.

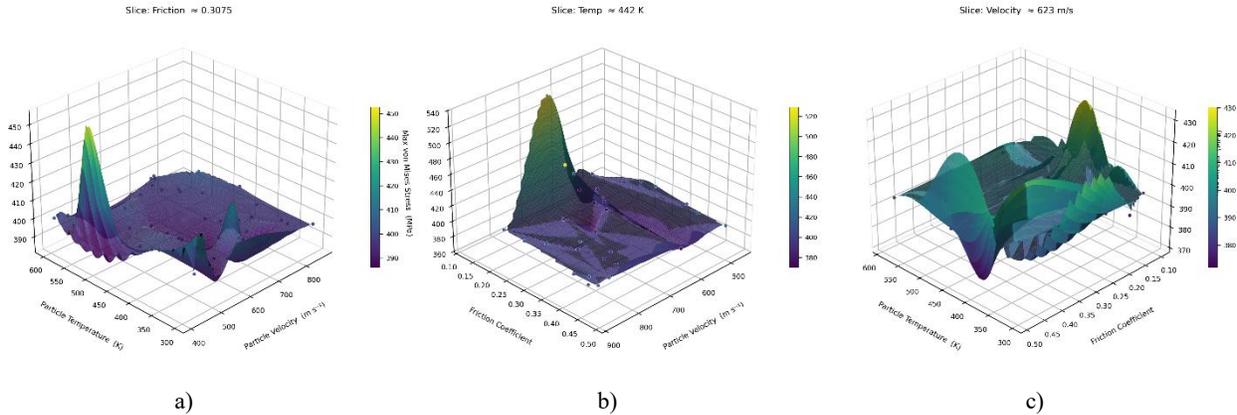

| a) | b) | c) |

**Figure 12.** Three-dimensional response surfaces of maximum von Mises stress (MPa) obtained by fixing one input parameter near its median value and sweeping the remaining two: (a) friction coefficient fixed at approximately 0.31, sweeping particle velocity and particle temperature; (b) particle temperature fixed at approximately 442 K, sweeping particle velocity and friction coefficient; and (c) particle velocity fixed at approximately 623 m s$^{-1}$, sweeping particle temperature and friction coefficient. Surfaces are generated via cubic interpolation on a structured meshgrid, with overlaid scatter points representing actual simulation samples coloured by their corresponding von Mises stress values. In contrast to the other output targets, all three panels exhibit pronounced localised peaks and irregular surface topography rather than smooth monotonic trends, reflecting the competing contributions of deformation hardening and thermal softening that produce a highly sensitive, non-monotonic stress response across the full input space.

Figure 13a (Velocity × Temperature) displays a smooth and well-structured response, with deformation ratio increasing steadily and monotonically with particle velocity while particle temperature contributes a comparatively minor modulating effect along the orthogonal axis. The broadly spaced and nearly vertical isolines across most of the parameter space confirm that velocity is the dominant control variable, with the transition from low to high deformation ratio occurring progressively and without abrupt discontinuities, suggesting that the geometric deformation of the particle upon impact is a relatively well-conditioned function of the kinetic energy input under typical cold spray operating conditions. Figure 13b (Velocity × Friction Coefficient) exhibits strong vertical banding analogous to the equivalent PEEQ projection, with isolines oriented almost entirely parallel to the friction axis and the deformation ratio gradient concentrated almost exclusively along the velocity direction. The wide colorbar range extending to approximately 1.24, driven by isolated high-value scatter points clustered at elevated velocities, reinforces that friction coefficient has a negligible systematic influence on particle deformation geometry and that the high-value outliers at the right boundary of the projection represent localised conditions rather than a generalised friction-driven trend. Figure 13c (Temperature × Friction Coefficient) presents the weakest and most featureless contour structure of the three panels, with the interpolated surface remaining nearly uniform at low deformation ratio values across the majority of the projection plane while isolated high-value scatter points are distributed without a coherent spatial pattern. The colorbar range expanding to approximately 1.68 is attributable entirely to boundary interpolation artefacts rather than physical variation, and the absence of any discernible gradient direction in this panel confirms that deformation ratio, more than any other output target in this study, is overwhelmingly determined by particle velocity alone and is largely insensitive to the combined influence of temperature and friction coefficient when velocity is excluded from consideration.

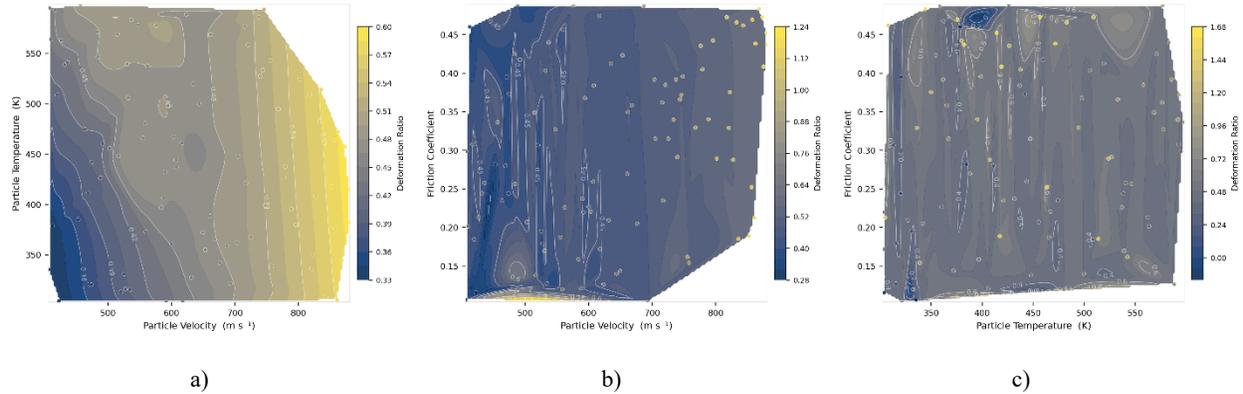

a)             b)             c)

**Figure 13.** Two-dimensional contour projections of deformation ratio across the three pairwise input combinations: (a) particle velocity versus particle temperature, (b) particle velocity versus friction coefficient, and (c) particle temperature versus friction coefficient. Filled contours represent the interpolated deformation ratio response surface obtained via cubic grid data interpolation, white isolines denote constant deformation boundaries, and overlaid scatter points correspond to actual simulation samples coloured by their respective deformation ratio values. The near-vertical isoline orientation and smooth gradient structure in (a) and (b) confirm that deformation ratio is overwhelmingly governed by particle velocity, while the structurally featureless surface in (c) demonstrates that temperature and friction coefficient carry negligible independent explanatory power for this output, making deformation ratio the most kinetically dominated and parametrically concentrated target among all five outputs considered in this study.

Figure 14a (Friction ≈ 0.31, Velocity × Temperature slice) produces the smoothest and most geometrically regular surface among all five output targets at this slice condition, displaying a broad, gently inclined plane that rises steadily with increasing particle velocity while remaining comparatively flat across the temperature axis. The excellent agreement between the interpolated mesh and the overlaid scatter points across the full sweep range, with minimal deviation or surface distortion, confirms that at a fixed friction coefficient the deformation ratio responds to velocity in a near-linear and thermally insensitive manner, consistent with its characterisation as the most kinetically dominated output in the dataset. Figure 14b (Temperature ≈ 442 K, Velocity × Friction slice) presents a thin, blade-like surface geometry that narrows sharply toward the low-velocity and low-friction corner of the parameter space, with the deformation ratio rising steadily along the velocity axis while remaining largely invariant across the friction dimension throughout the majority of the swept range. The surface maintains good coherence and close agreement with the scatter points, further reinforcing that friction coefficient exerts no meaningful systematic influence on particle deformation geometry and that the response at fixed temperature is effectively a univariate function of velocity with minimal secondary modulation. Figure 14c (Velocity ≈ 623 m s$^{-1}$, Temperature × Friction slice) departs from the regularity of the preceding two panels, exhibiting a broadly elevated plateau punctuated by a pronounced central depression and sharp edge corrugation concentrated in the low-temperature and low-friction corner of the plane.

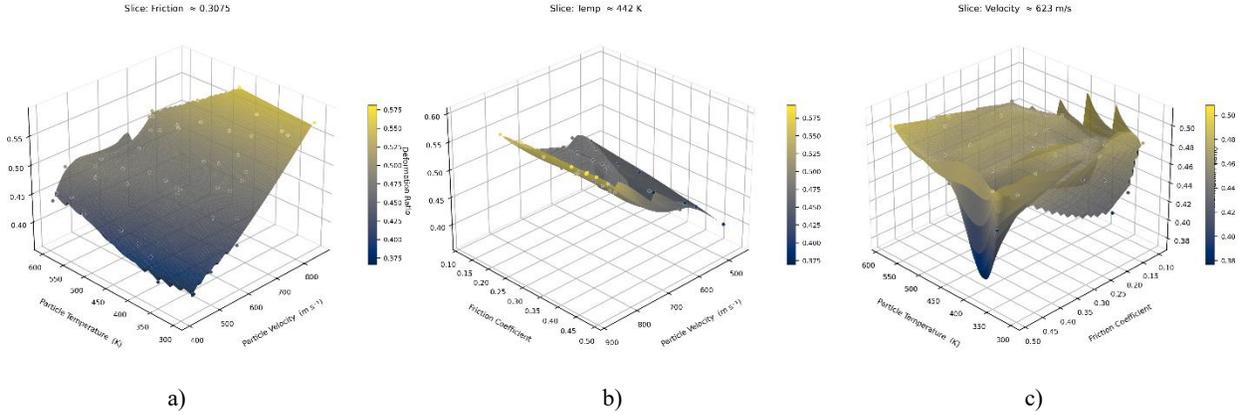

**Figure 14.** Three-dimensional response surfaces of deformation ratio obtained by fixing one input parameter near its median value and sweeping the remaining two: (a) friction coefficient fixed at approximately 0.31, sweeping particle velocity and particle temperature; (b) particle temperature fixed at approximately 442 K, sweeping particle velocity and friction coefficient; and (c) particle velocity fixed at approximately 623 m s$^{-1}$, sweeping particle temperature and friction coefficient. Surfaces are generated via cubic interpolation on a structured meshgrid, with overlaid scatter points representing actual simulation samples coloured by their corresponding deformation ratio values. The smooth, near-planar geometry of panels (a) and (b) reflects the dominant and largely linear influence of particle velocity on geometric deformation, while the structured plateau and localised depression visible in panel (c) reveal a residual coupled sensitivity to particle temperature and friction coefficient at intermediate velocities, confirming that a full three-dimensional input representation remains necessary even for this comparatively well-conditioned output target.

### 3.3. Evaluation of Geometric Deep Learning Models

Tables 3-6 and Figure 15 present the prediction performance of the four geometric deep learning algorithms across all output targets. For maximum equivalent plastic strain (Table 3) and average contact plastic strain (Table 4), both GraphSAGE and GAT achieve consistently high $R^2$ values exceeding 0.93, with GAT attaining the best overall performance at $R^2$ = 0.9700 and 0.9304 respectively, confirming that spatial graph-based message passing is well-suited to capturing the localised deformation response. For maximum temperature (Table 5), GraphSAGE and GAT again dominate with $R^2$ values of 0.9594 and 0.9465, while TDA-MLP produces a negative $R^2$ of −0.2104 indicating that topological feature augmentation alone is insufficient to resolve the thermal field without explicit graph connectivity. For particle deformation ratio (Table 6), GraphSAGE achieves the highest $R^2$ of 0.9360 closely followed by GAT at 0.8867, whereas ChebSpectral and TDA-MLP perform considerably worse across all targets suggesting that the Chebyshev spectral approximation and the persistence-homology-augmented MLP are less effective at generalising the cold spray input-output relationships compared to the spatial neighbourhood aggregation strategies employed by GraphSAGE and GAT.

**Table 3.** Prediction performance for maximum equivalent plastic strain (max_peeq)

| Algorithm | MSE | MAE | $R^2$ |
|---|---|---|---|
| GraphSAGE | 0.069657 | 0.185068 | 0.9414 |
| ChebSpectral | 0.625710 | 0.603719 | 0.4734 |
| TDA-MLP | 0.046952 | 0.170959 | 0.9605 |
| GAT | 0.035597 | 0.140189 | 0.9700 |

**Table 4.** Prediction performance for average contact plastic strain (avg_peeq_contact)

| Algorithm | MSE | MAE | $R^2$ |
|---|---|---|---|
| GraphSAGE | 0.000028 | 0.004320 | 0.9821 |

| Algorithm | MSE | MAE | $R^2$ |
|---|---|---|---|
| ChebSpectral | 0.000774 | 0.023146 | 0.5062 |
| TDA-MLP | 0.001676 | 0.033979 | -0.0688 |
| GAT | 0.000109 | 0.007314 | 0.9304 |

Table 5. Prediction performance for maximum temperature (max_temp_K)

| Algorithm | MSE | MAE | $R^2$ |
|---|---|---|---|
| GraphSAGE | 283.043396 | 12.343616 | 0.9594 |
| ChebSpectral | 4642.883301 | 52.834316 | 0.3335 |
| TDA-MLP | 8430.983398 | 72.116943 | -0.2104 |
| GAT | 372.935272 | 14.187658 | 0.9465 |

Table 6. Prediction performance for particle deformation ratio

| Algorithm | MSE | MAE | $R^2$ |
|---|---|---|---|
| GraphSAGE | 0.000325 | 0.012830 | 0.9360 |
| ChebSpectral | 0.003503 | 0.038748 | 0.3111 |
| TDA-MLP | 0.005781 | 0.059728 | -0.1367 |
| GAT | 0.000576 | 0.016186 | 0.8867 |

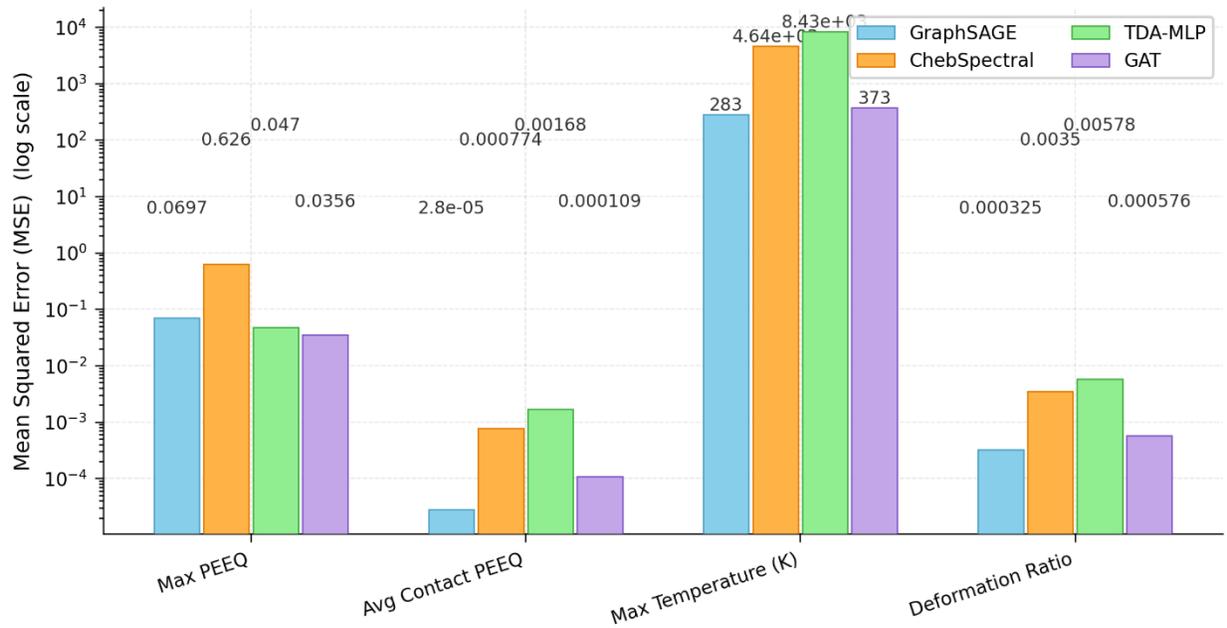

a)

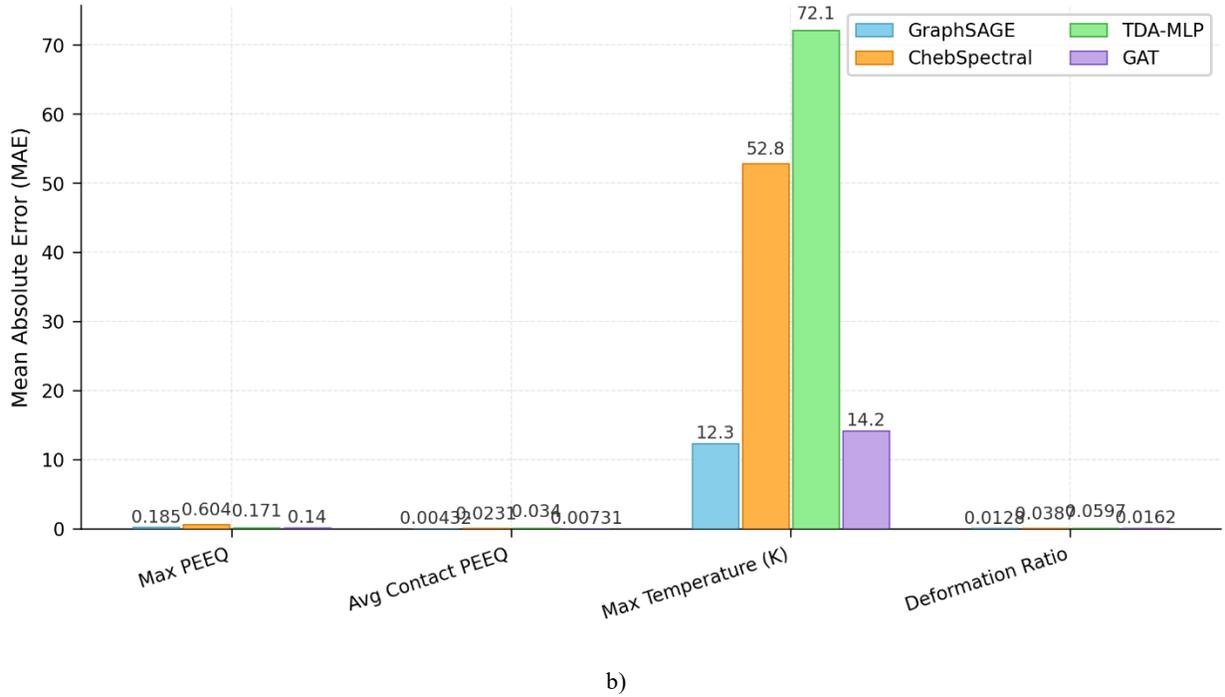

b)

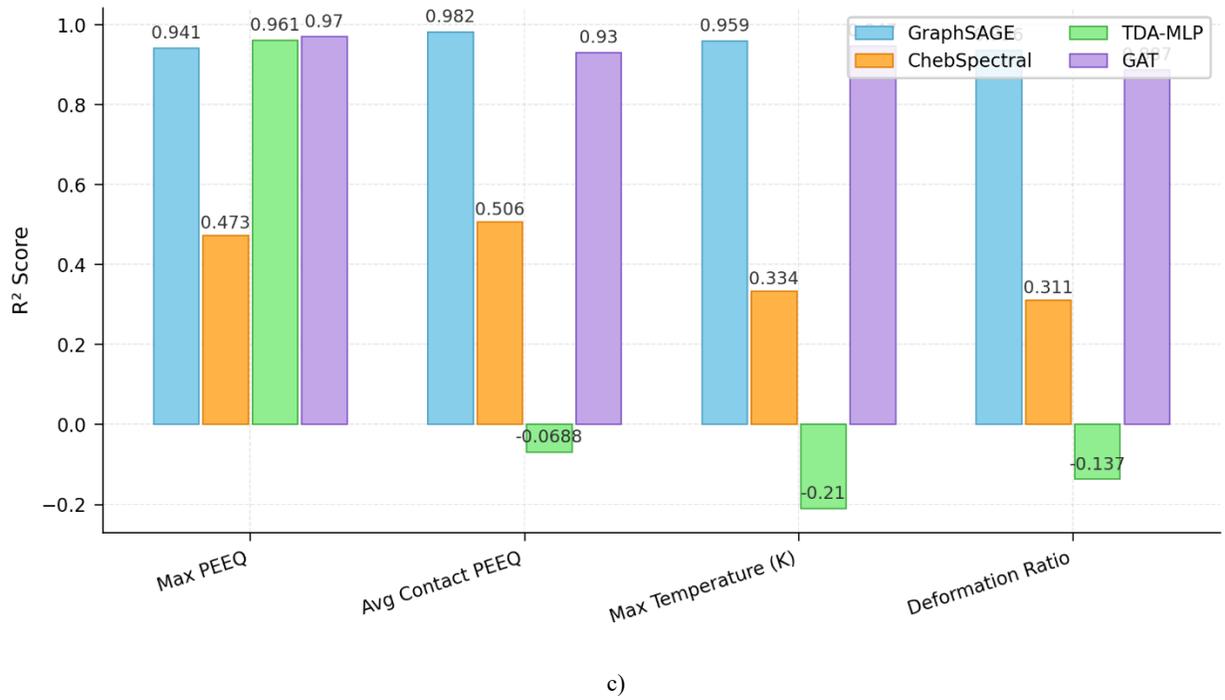

c)

**Figure 15.** Prediction performance of the four geometric deep learning algorithms across all output targets: (a) mean squared error (MSE, log scale), (b) mean absolute error (MAE), and (c) R² score. GraphSAGE and GAT consistently achieve the lowest error and highest R² values across all targets, while ChebSpectral and TDA-MLP perform considerably worse, with TDA-MLP yielding negative R² for multiple outputs, confirming the superiority of spatial graph-based aggregation approaches for cold spray process prediction.

Figure 16 presents the predicted versus actual plots for maximum equivalent plastic strain (max PEEQ). GraphSAGE and GAT produce tight point clusters along the ideal diagonal with minimal scatter, consistent with their $R^2$ values of 0.94 and 0.97, while ChebSpectral shows considerable dispersion and TDA-MLP displays a moderately improved but still less precise fit. Figure 17 shows the equivalent plots for average contact plastic strain (avg PEEQ). GraphSAGE achieves the closest alignment to the identity line with $R^2 = 0.98$, while GAT follows closely at 0.93. The TDA-MLP scatter plot reveals a near-random distribution of points with no discernible trend, consistent with its negative $R^2$ of −0.07, indicating a complete failure to learn the underlying relationship for this target. Figure 18 presents the predicted versus actual plots for maximum temperature. GraphSAGE and GAT again demonstrate strong diagonal alignment with low scatter, reflecting their $R^2$ values above 0.94, whereas ChebSpectral exhibits a broadly dispersed cloud and TDA-MLP produces the most severely degraded fit with points distributed almost randomly, consistent with its negative $R^2$ of −0.21 and an MAE exceeding 72 K. Figure 19 displays the predicted versus actual plots for particle deformation ratio. GraphSAGE achieves the tightest fit at $R^2 = 0.936$, with GAT following at 0.887, both showing well-concentrated point distributions close to the diagonal. ChebSpectral and TDA-MLP scatter plots deviate substantially from the identity line, with TDA-MLP again producing a negative $R^2$ of −0.137, reinforcing the consistent pattern observed across all four targets whereby spatial graph aggregation methods generalise effectively while spectral and topological augmentation approaches fail to capture the deformation geometry with comparable reliability.

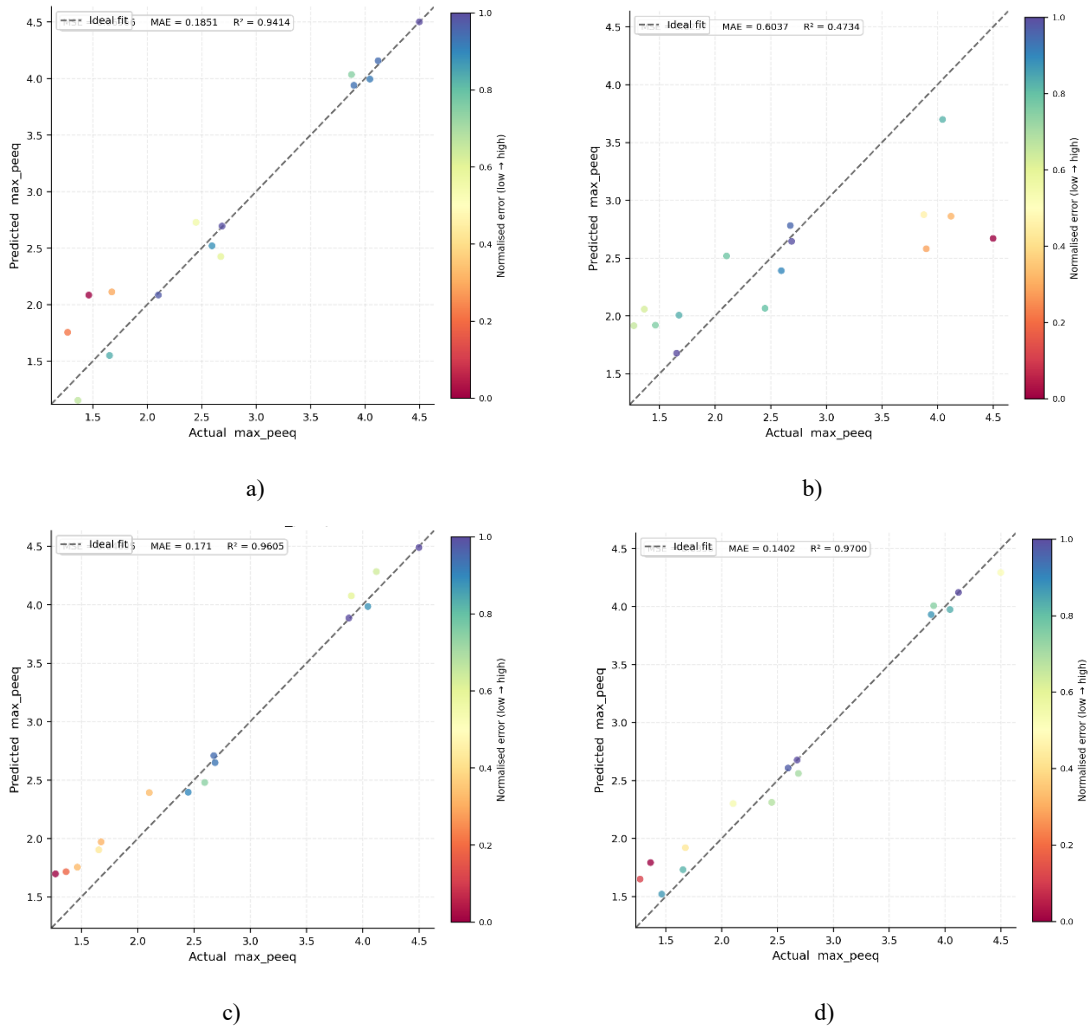

a)

b)

c)

d)

**Figure 16.** Plots for actual vs predicted values of max_peeq obtained using (a) GraphSAGE, (b) ChebSpectral, (c) TDA-MLP, and (d) GAT.

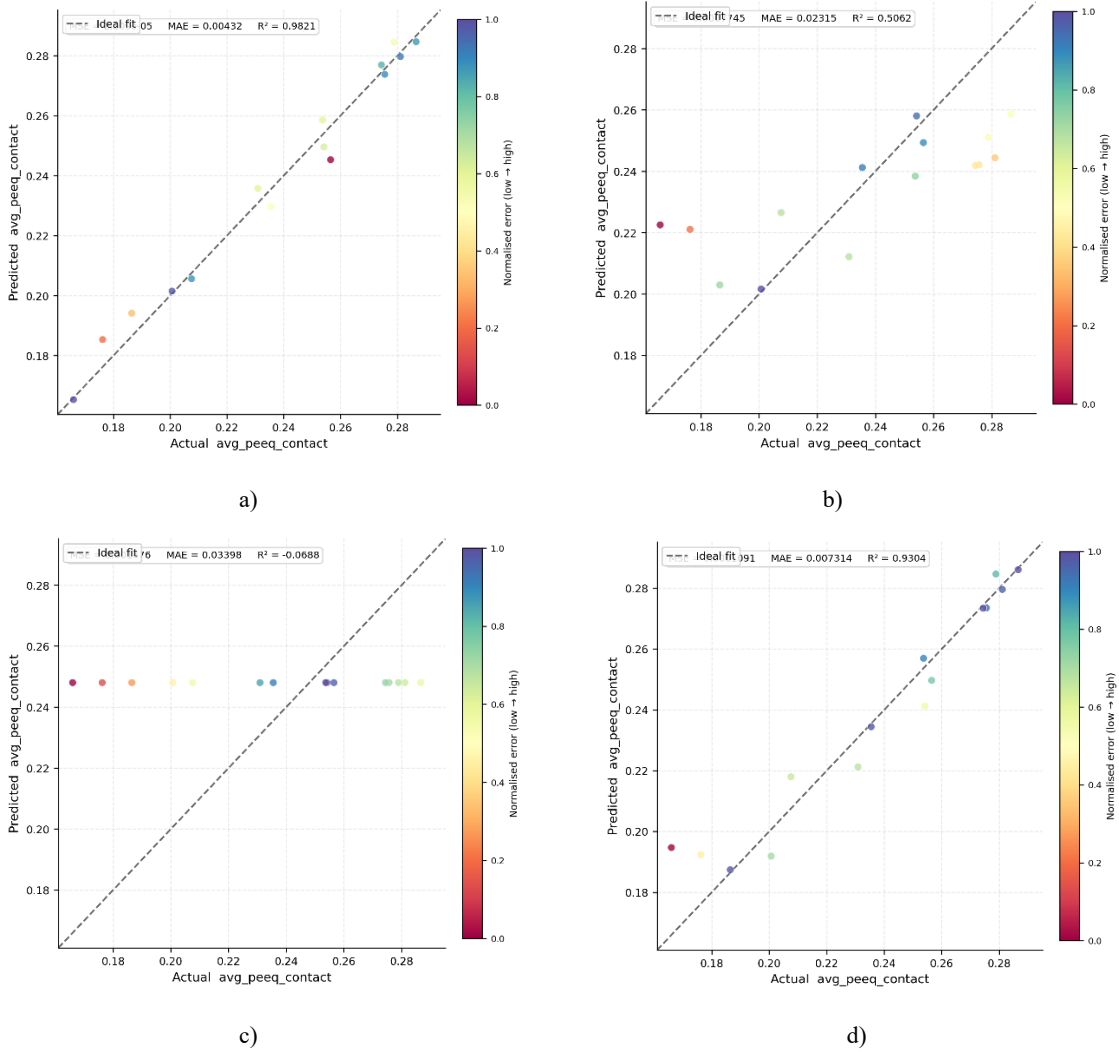

a)

b)

c)

d)

**Figure 17.** Plots for actual vs predicted values of avg_peeq obtained using (a) GraphSAGE, (b) ChebSpectral, (c) TDA-MLP, and (d) GAT.

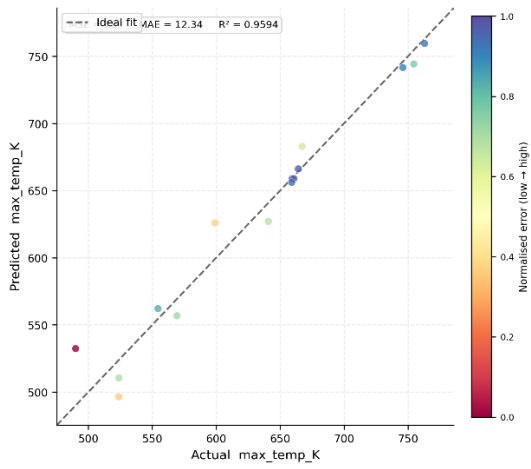
a)

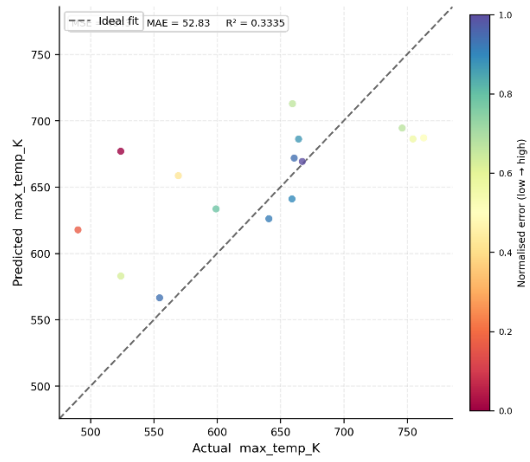
b)

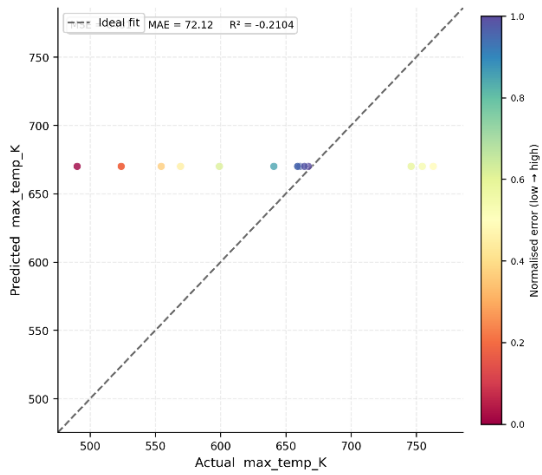
c)

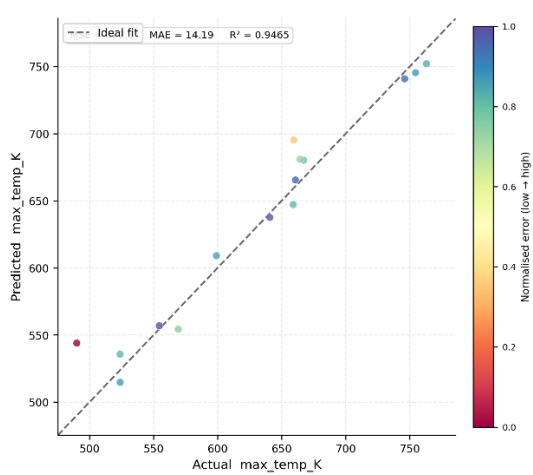
d)

**Figure 18.** Plots for actual vs predicted values of maximum temperature obtained using (a) GraphSAGE, (b) ChebSpectral, (c) TDA-MLP, and (d) GAT.

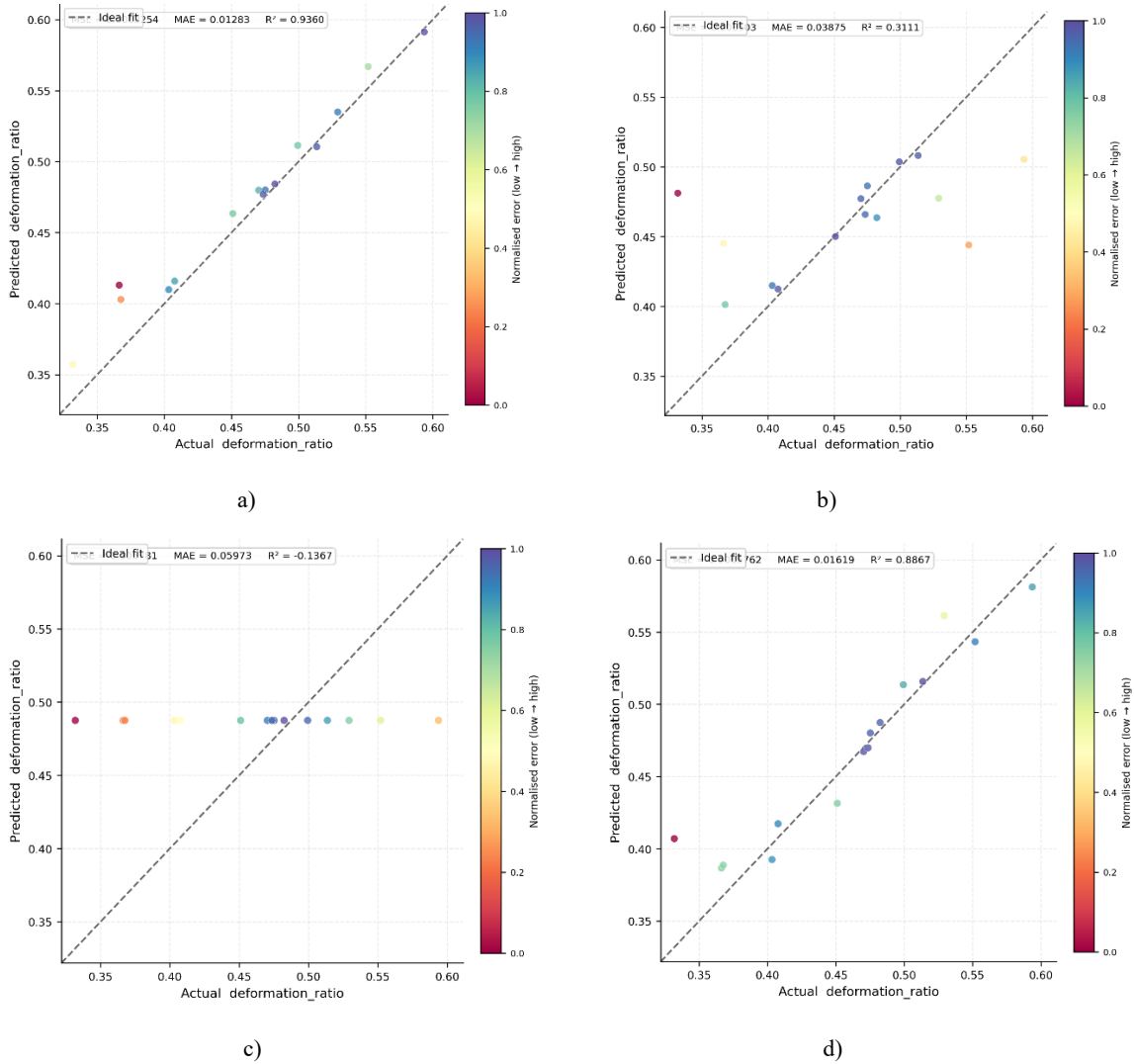

**Figure 19.** Plots for actual vs predicted values of deformation ratio obtained using (a) GraphSAGE, (b) ChebSpectral, (c) TDA-MLP, and (d) GAT.

## 4. Conclusion

This study developed and evaluated four geometric deep learning architectures as surrogate models for cold spray particle impact prediction, trained on a parametric finite element simulation dataset spanning particle velocity, temperature, and friction coefficient. The construction of a k-nearest-neighbour graph in the input feature space allowed each model to leverage spatial similarity between process conditions, providing a physically motivated inductive bias that conventional point-wise architectures cannot exploit. Three-dimensional feature space analysis confirmed that particle velocity is the dominant governing parameter across most output targets, with temperature and friction coefficient contributing secondary and coupled influences that necessitate a full three-dimensional input representation.

GraphSAGE and the geometric attention network consistently achieved R² values exceeding 0.93 across maximum plastic strain, average contact plastic strain, maximum temperature, and deformation ratio, with GAT attaining peak accuracy of R² = 0.97 for maximum equivalent plastic strain. The Chebyshev spectral network and the topological data analysis augmented multilayer perceptron performed considerably worse, yielding negative R² values for several targets and indicating that spectral and topological augmentation strategies alone are insufficient to resolve the non-linear response surfaces of cold spray impact. These findings establish spatial graph-based neighbourhood aggregation as a robust and computationally efficient surrogate modelling strategy for cold spray process optimisation, and provide a foundation for extending the geometric deep learning framework to multi-particle deposition scenarios and microstructure-level property prediction in future work.